\documentclass[conference]{IEEEtran}
\usepackage{graphicx}
\usepackage{amsmath, amssymb, amsfonts}
\usepackage{algorithm}
\usepackage{algpseudocode}
\usepackage{tikz}
\usepackage{pgfplots}
\pgfplotsset{compat=1.18}
\usepgfplotslibrary{groupplots, fillbetween}
\usetikzlibrary{arrows.meta, patterns, calc}
\usepackage{subcaption}
\usepackage{caption}
\usepackage{multirow}
\usepackage{listings}
\usepackage{booktabs}
\usepackage{url}
\usepackage{hyperref}
\usepackage{booktabs}
\usepackage{multirow}
\usepackage{longtable}
\usepackage{pdflscape}
\usepackage{array}

\ifCLASSINFOpdf
\else
\fi
\begin{document}
%
% paper title
% Titles are generally capitalized except for words such as a, an, and, as,
% at, but, by, for, in, nor, of, on, or, the, to and up, which are usually
% not capitalized unless they are the first or last word of the title.
% Linebreaks \\ can be used within to get better formatting as desired.
% Do not put math or special symbols in the title.
\title{Label-Efficient Time Series Classification at Scale: A Dual-Stream OSSE-LSTM with Counterfactual Attribution}

%\author{\IEEEauthorblockN{Anonymous}}

% author names and affiliations
% use a multiple column layout for up to three different
% affiliations

\author{\IEEEauthorblockN{Nguyen Ho$^{*}$, Bach Tung Tran$^{**}$, Long Van Ho$^{**}$, Trung Ky Nguyen$^{**}$, Zhenchang Xia$^{++}$, Bolong Zheng$^{++}$}
\IEEEauthorblockA{* Loyola University Maryland, USA \\
** School of Computer Science and Engineering, International University, Vietnam\\
++ School of Compute Science and Artificial Intelligence, Wuhan University of Technology, China \\
Email: tnho@loyola.edu, \{ITITIU21340, hlvan, ntky\}@hcmiu.edu.vn, \{zcxia.cs, bolongzheng\}@whut.edu.cn}}
%\and
%\IEEEauthorblockN{Homer Simpson}
%\IEEEauthorblockA{Twentieth Century Fox\\
%Springfield, USA\\
%Email: homer@thesimpsons.com}
%\and
%\IEEEauthorblockN{James Kirk\\ and Montgomery Scott}
%\IEEEauthorblockA{Starfleet Academy\\
%San Francisco, California 96678--2391\\
%Telephone: (800) 555--1212\\
%Fax: (888) 555--1212}}

% conference papers do not typically use \thanks and this command
% is locked out in conference mode. If really needed, such as for
% the acknowledgment of grants, issue a \IEEEoverridecommandlockouts
% after \documentclass

% for over three affiliations, or if they all won't fit within the width
% of the page, use this alternative format:
% 
%\author{\IEEEauthorblockN{Michael Shell\IEEEauthorrefmark{1},
%Homer Simpson\IEEEauthorrefmark{2},
%James Kirk\IEEEauthorrefmark{3}, 
%Montgomery Scott\IEEEauthorrefmark{3} and
%Eldon Tyrell\IEEEauthorrefmark{4}}
%\IEEEauthorblockA{\IEEEauthorrefmark{1}School of Electrical and Computer Engineering\\
%Georgia Institute of Technology,
%Atlanta, Georgia 30332--0250\\ Email: see http://www.michaelshell.org/contact.html}
%\IEEEauthorblockA{\IEEEauthorrefmark{2}Twentieth Century Fox, Springfield, USA\\
%Email: homer@thesimpsons.com}
%\IEEEauthorblockA{\IEEEauthorrefmark{3}Starfleet Academy, San Francisco, California 96678-2391\\
%Telephone: (800) 555--1212, Fax: (888) 555--1212}
%\IEEEauthorblockA{\IEEEauthorrefmark{4}Tyrell Inc., 123 Replicant Street, Los Angeles, California 90210--4321}}

% use for special paper notices
%\IEEEspecialpapernotice{(Invited Paper)}

% make the title area
\maketitle

% As a general rule, do not put math, special symbols or citations
% in the abstract
\begin{abstract}

Time series are produced continuously at enormous scale by industrial equipment, wearables, power grids, and clinical monitors, yet annotation remains manual, expensive, and expert-dependent. The binding constraint in large-scale time series analytics is therefore not data volume but \emph{label} volume, and
the question facing a practitioner is concrete: how many examples per class must be labeled before a classifier becomes usable? We study this question directly, in a regime where the label space is fixed and known in advance and the decision rule must be constructed from only $K$ labeled examples per class. We propose
\textbf{Dual-Stream OSSE-LSTM}, an episodic metric-learning framework that pairs an Omni-Scale CNN with Squeeze-and-Excitation recalibration, for multi-scale motif extraction without per-dataset kernel tuning, with a Bidirectional LSTM for global temporal context. The two streams are independently normalized and fused into a prototype-oriented embedding. Because decisions taken from a few labels must also be explainable, we introduce \textbf{Counterfactual Integrated Gradients (C-IG)}, which attributes the prototype margin between target and opposing classes rather than an isolated classifier logit, and reuses the resulting maps as soft masks for test-time prototype refinement without updating the encoder.On 19 univariate UCR datasets, OSSE-LSTM attains the highest average accuracy and
per-dataset win count at every support size from $K=2$ to $K=8$, and its accuracy remains within a $0.36$-point band ($96.36$-$96.72\%$) across that range. Its weakest configuration still exceeding the best result any compared baseline achieves at any $K$ ($93.99\%$). Further, the baselines show no comparable stability: convolutional models lose up to $12$ points as $K$ grows, since additional support samples displace rather than stabilize their prototypes, while self-supervised representations improve with $K$ yet remain far below our method. Further, our C-IG-guided refinement delivers consistent gains where unguided noise augmentation causes catastrophic degradation.

%Few-shot time series classification is challenging because limited labeled data makes deep models prone to overfitting, while discriminative patterns may appear as both local motifs and long-range temporal dependencies. We propose Dual-Stream OSSE-LSTM, an episodic metric-learning framework that combines an Omni-Scale CNN with Squeeze-and-Excitation blocks for multi-scale motif extraction and a Bidirectional LSTM for temporal context modeling. The two feature streams are separately normalized and concatenated before prototype-based classification.

%To improve interpretability, we introduce Counterfactual Integrated Gradients (C-IG), which explains decisions by integrating gradients between target and opposing class representatives in the prototypical space. The resulting attribution maps are also used to guide test-time augmentation: background regions are softly perturbed while discriminative regions are preserved, enabling more stable prototype estimation during inference. Experiments on 18 UCR and 9 UEA datasets show that OSSE-LSTM performs consistently under limited-data settings, achieving 92.79\% average accuracy on univariate datasets at various training-data regime.

\vspace{0.5cm}
\end{abstract}

% For peer review papers, you can put extra information on the cover
% page as needed:
% \ifCLASSOPTIONpeerreview
% \begin{center} \bfseries EDICS Category: 3-BBND \end{center}
% \fi
%
% For peerreview papers, this IEEEtran command inserts a page break and
% creates the second title. It will be ignored for other modes.
\IEEEpeerreviewmaketitle

\section{Introduction}

Modern infrastructure emits time series continuously. A single industrial plant
instruments thousands of channels at sub-second resolution; consumer wearables
stream physiological signals from hundreds of millions of devices; smart meters,
vehicle telemetry, and clinical monitors each generate archives that grow without
bound. Collecting this data is now inexpensive, but interpreting it is not. Deep
neural networks achieve strong time series classification (TSC) performance by
learning representations directly from raw signals \cite{mohammadi2024deep}, but
typically require abundant labeled data, an assumption that fails precisely
where these archives are largest. Labeling a heartbeat as an infarction, a vibration trace as a bearing wear, or a power signature as an appliance class requires a domain expert, a controlled experiment, or a confirmed downstream outcome. None of these scales with sampling frequency, as data accumulates at machine speed, while labels accumulate at human speed.

This asymmetry reframes what scalability means for time series analytics. The scarce resource is no longer storage or compute, but \emph{annotation budget}, and the governing question is operational rather than asymptotic: how many examples per class must an expert label before the resulting classifier is fit for deployment? This question motivates our work. We study TSC in the \textbf{label-efficient regime}, where a classifier must label data accurately from only $K$ examples per class over a fixed, known label space, the situation of an analyst who can annotate two or three representative instances of each condition, but not a corpus. Our aim is to propagate a small vocabulary of expert labels across a large body of unlabeled signals, using episodic $K$-shot training and prototype-based inference. %, and we treat $K$ as the primary experimental variable because it is the quantity a practitioner actually controls.

Metric-based learning is a natural fit for this label-efficient regime. For instance, Prototypical Networks \cite{snell2017prototypical} classify a query by comparing its embedding to class
prototypes computed from a small support set, avoiding a separate classifier per
task. %revising the labeled set requires recomputing a mean rather than retraining a classifier head, which is an appealing property when an archive is annotated incrementally. 
Time series data, however, complicate prototype
estimation in two ways that become more acute at scale. First, \textbf{variety}: discriminative
structure appears at radically different temporal scales and in different forms
across domains, e.g., a millisecond ECG deflection, a multi-second gait cycle, a slow
drift in a control trace. An encoder tuned to one morphology transfers poorly to
another, and hand-tuning receptive fields per collection does not scale to
archives containing thousands of heterogeneous signal types. Second,
\textbf{veracity}: when a prototype is the mean of two or three examples,
background noise, baseline drift, and phase shift are not averaged away, so a
prototype can be displaced by variation that carries no class information. This
second difficulty has a consequence that is easy to overlook, and that our
experiments make explicit: if an embedding is not organized around prototype
geometry, enlarging the support set will not help. Additional labeled examples
introduce additional intra-class variation, and a prototype computed over them
can drift further from the class structure than one computed over fewer. Label
efficiency is therefore not simply a matter of tolerating small $K$, but it is a
matter of ensuring that each labeled example contributes usefully to the
prototype.

Existing encoders address only part of this. Convolutional and shapelet-based
models extract local patterns effectively \cite{wang2017time,tang2020interpretable}
but often rely on temporal pooling that weakens global ordering. Recurrent models
such as LSTM capture sequential dependencies \cite{hochreiter1997long} yet overfit
under very limited training data. Hybrid CNN--LSTM models exist
\cite{karim2017lstm} but are designed for standard supervised learning rather than
few-shot training.

Interpretability is a further requirement that is important in label-efficient regime. When a decision rests on two labeled examples, an operator is entitled to
ask what evidence supports it, and in healthcare and industrial monitoring, an
unexplained prediction is often unusable regardless of its accuracy. Existing work such as Integrated
Gradients (IG) \cite{sundararajan2017axiomatic} attributes predictions by
integrating gradients from a baseline to the input, but the baseline choice is
non-trivial for time series, where a zero signal is not a realistic reference
\cite{sturmfels2020visualizing}. Furthermore, like most attribution methods, IG explains a single class score in isolation. Prototype-based
classification, however, is inherently contrastive: a query is assigned to a class
because it is closer to that prototype than to competing ones, so a useful
explanation should highlight regions that drive this separation rather than
regions that simply activate one class.

To address these challenges, we propose \textbf{Dual-Stream OSSE-LSTM}, an
episodic metric-learning framework for label-efficient TSC. It combines two
complementary encoders: an Omni-Scale CNN with Squeeze-and-Excitation blocks for
multi-scale motif extraction \cite{tang2020omni,hu2018squeeze}, which covers a
broad range of receptive-field lengths by construction and therefore removes
per-dataset kernel-size tuning, and a Bidirectional LSTM for global temporal
context, which preserves the ordering information that pooling discards. Their
outputs are independently normalized and concatenated for prototype-based
classification. We further introduce \textbf{Counterfactual Integrated Gradients
(C-IG)}, an attribution module for prototypical decisions that integrates
gradients along paths between target and opposing class representatives, rather
than from an arbitrary zero baseline, and thereby attributes the prototype margin
that the model actually uses to decide. Beyond visualization, C-IG maps guide
test-time prototype refinement: low-attribution background regions of support
samples are softly perturbed while high-attribution regions are preserved, and
original and augmented embeddings are averaged for more stable prototypes at
inference. Because this refinement leaves the encoder untouched, it costs nothing
in training when the labeled set is revised.
The main contributions of our paper are:
\begin{itemize}
    \item \textbf{A label-efficient formulation of large-scale TSC:} We formalize time series
    classification problem over a fixed, known label space from $K$ labeled examples per
    class. This is the regime that governs annotation cost in large operational time series archives. %and we evaluate it along the axis a practitioner controls, sweeping the support size $K$ and, secondarily, the labeled training fraction $\rho$.

    \item \textbf{Dual-Stream OSSE-LSTM:} We propose a dual-stream encoder for
    label-efficient time series classification that integrates omni-scale
    convolutional representations for multi-scale local temporal patterns with
    bidirectional LSTM representations for temporal context. The two streams are
    independently normalized and fused into a prototype-oriented embedding space
    suitable for metric-based classification under limited supervision, and the
    omni-scale design covers heterogeneous morphologies without dataset-specific
    receptive-field tuning, making a single configuration viable
    across a diverse archive.

    \item \textbf{Counterfactual Integrated Gradients (C-IG):} We propose a prototype-aware
    attribution method that interprets metric-based decisions by attributing the
    prototype margin between target and opposing classes, rather than a fixed
    classifier logit. Beyond visualization, C-IG attribution maps serve as soft
    masks for test-time support augmentation, enabling prototype refinement
    without updating the trained encoder.

    \item \textbf{An annotation-budget evaluation on 19 univariate UCR datasets.}
    Against the representative baselines, our OSSE-LSTM attains the highest average accuracy and
    per-dataset win count at every support size tested. It reaches $96.36\%$ from
    two labeled examples per class, exceeding every baseline's eight-shot
    result, and varies by under $0.4$ points from $2$- to $8$-shot, whereas
    convolutional baselines lose up to $12$ points over the same range. Further, C-IG-guided refinement yields consistent gains where unguided augmentation degrades
    accuracy severely.
\end{itemize}

\section{Related Work}
\label{sec:related_work}

\subsection{Time Series Representation Learning}

Classical TSC methods \cite{ho2016adaptive, ho2019amic, ho2019efficient, ho2020efficient, ho2020efficient1, ho2021efficient} include distance-based, dictionary-based, interval-based, and shapelet-based approaches. Among them, Dynamic Time Warping is widely used to handle temporal misalignment, while shapelet methods identify discriminative subsequences that are useful for both classification and interpretation \cite{bagnall2017great,ye2009time, ho2012activity, gribaudo2014analysis, barkat2014open}. These methods provide strong inductive bias, but they often rely on hand-crafted distances or expensive search procedures.

Deep learning methods \cite{van2022efficient, ho2022unified, ho2023mining, tran2023exploring, hoang2023experimental} reduce the need for manual feature design by learning temporal representations from raw data. For example, Convolutional Neural Networks and ResNet-style models \cite{wang2017time} demonstrated the effectiveness of 1D convolutions for TSC, and InceptionTime \cite{ismail2020inceptiontime} further improved multi-scale convolutional modeling through inception-style modules and ensembles. Similarly, OS-CNN \cite{tang2020omni} follows a related direction by using multiple prime-numbered kernel sizes to cover different temporal scales. These convolutional models are effective at extracting local patterns, but the global temporal dependencies may be weak after pooling.

Hybrid architectures integrate convolutional and recurrent components to jointly capture local temporal patterns and long-range sequential dependencies \cite{ho2023causal, ho2024mining, long2024efficient, ho2024mining1, ngo2025data, ho2025efficient, ho2026distributed}. Representative models such as LSTM-FCN and its multivariate extensions process time series through parallel convolutional and recurrent branches; Squeeze-and-Excitation (SE) blocks further refine the convolutional pathway by adaptively recalibrating the responses of the feature channel \cite{hu2018squeeze}. We advance this paradigm by situating hybrid temporal representations within an episodic metric-learning framework, so that reliable prototypes can be formed from a few labeled examples per class rather than from a fully annotated training set.

\subsection{Few-Shot Time Series Classification}

Few-shot learning aims to build classifiers that generalize from a small support set \cite{cappiello2015co, ho2015data, ho2016characterizing, thi2017towards, ho2017improving}. Metric-based methods are particularly suited to this setting, as they learn an embedding space in which classification reduces to nearest-prototype comparison. Prototypical Networks \cite{snell2017prototypical}, for instance, represent each class by the mean embedding of its support examples and classify queries by distance to these prototypes.

Several works have adapted the prototype idea to time series. TapNet \cite{zhang2020tapnet} combines convolutional encoders with attentional prototypes for multivariate TSC. Self-supervised approaches such as TS2Vec \cite{yue2022ts2vec} learn transferable temporal representations that transfer to limited-label settings. DPSN \cite{tang2020interpretable} couples shapelet learning with prototypical classification, highlighting the value of local discriminative subsequences. However, existing methods typically emphasize either local motifs or global sequence embeddings. Our OSSE-LSTM is designed to integrate both within a single framework.

\subsection{Interpretability for Prototype-Based Time Series Models}

Interpretability is another concern in TSC, where predictions must be explainable to end-users. Gradient- and embedding-based explanations are common: Grad-CAM-style methods \cite{selvaraju2017grad} highlight discriminative regions via gradients flowing into the final convolutional layer, but their resolution is inherently limited by that of the final feature maps. Integrated Gradients \cite{sundararajan2017axiomatic} offers finer feature-level attribution, yet depends on a baseline choice that is non-trivial for time series, where a zero signal is not a realistic reference \cite{sturmfels2020visualizing}. Gradient-based explanations can also be visually noisy in deep networks \cite{smilkov2017smoothgrad}, a behavior linked to gradient instability and shattered gradients during training \cite{balduzzi2017shattered}. 

For prototype-based networks \cite{snell2017prototypical}, explanations should reflect the metric decision rule: a sample is classified not by an isolated logit, but by its relative distance to competing class prototypes. Building on the broader idea of contrastive explanations \cite{dhurandhar2018explanations}, we propose Counterfactual Integrated Gradients (C-IG), introduced in Section \ref{sec:methodology}, which adopts this contrastive view by integrating gradients between target and opposing class representatives, aligning attribution with prototype separation. The resulting maps also serve an operational role, guiding test-time augmentation of support samples to make prototype estimation less sensitive to background noise.
\section{Preliminaries}
\label{sec:preliminaries}

This section defines the notation and core mathematical components used in our framework. %We first formulate episodic few-shot time series classification, then introduce prototypical metric learning, omni-scale receptive fields, channel recalibration, bidirectional recurrent temporal modeling, and integrated-gradient attribution. These definitions are used later in Section~\ref{sec:methodology}.

\subsection{Time Series and Episodic Few-Shot Classification}

Let $\mathcal{D}=\{(\mathbf{x}_i, y_i)\}_{i=1}^{n}$ be a labeled time series dataset of $n$ samples. Each sample $\mathbf{x}_i \in \mathbb{R}^{C \times L}$ has $C$ channels and $L$ time steps, with $C=1$ for univariate and $C>1$ for multivariate series. Each label $y_i \in \mathcal{Y} = \{1, 2, \ldots, M\}$, where $M$ is the number of classes.

Few-shot classification is evaluated through a sequence of episodes. In each \emph{episode}, a subset of $N$ classes $\mathcal{Y}_e \subseteq \mathcal{Y}$ is sampled, and for each class $c \in \mathcal{Y}_e$ we draw $K$ labeled \emph{support examples} and $R$ \emph{query examples}. The episode support and query sets are denoted as
\begin{equation}
    \mathcal{S} = \{(\mathbf{x}^{s}_{k,i}, c_k)\}_{k=1, i=1}^{N, K}
    \qquad
    \mathcal{Q} = \{(\mathbf{x}^{q}_{k,j}, c_k)\}_{k=1, j=1}^{N, R}
\end{equation}
where $k$ indexes classes in $\mathcal{Y}_e$, and $i, j$ index the support and query examples of class $c_k$. The superscript $s$ indicates a support sample, and $q$ indicates a query sample. The model must classify each query in $\mathcal{Q}$ using only $\mathcal{S}$. We refer to this as an $N$-way $K$-shot task.

%Few-shot classification is evaluated through a sequence of episodes. In each episode, a subset of $N$ classes $\mathcal{Y}_e=\{c_1^e,c_2^e,\ldots,c_N^e\}\subseteq\mathcal{Y}$ is sampled. For each selected class $c_k^e\in\mathcal{Y}_e$, the episode contains $K$ labeled support examples and $R$ query examples. We denote the class-specific support and query sets as follows:
%\begin{equation}
%    \mathcal{S}_{c_k^e}
%    =
%    \{(\mathbf{x}^{s}_{k,i},c_k^e)\}_{i=1}^{K}
%\end{equation}
%\begin{equation}
%    \mathcal{Q}_{c_k^e}
%    =
%    \{(\mathbf{x}^{q}_{k,j},c_k^e)\}_{j=1}^{R}
%\end{equation}

%Here, the superscript $s$ indicates a support sample, while the superscript $q$ indicates a query sample. The index $k$ identifies the selected class within the current episode, and $i$ and $j$ index the support and query examples of that class, respectively.

%The complete support set and query set of the episode are defined as:
%\begin{equation}
%    \mathcal{S}
%    =
%    \bigcup_{c_k^e\in\mathcal{Y}_e}
%    \mathcal{S}_{c_k^e}
%\end{equation}
%\begin{equation}
%    \mathcal{Q}
%    =
%    \bigcup_{c_k^e\in\mathcal{Y}_e}
%    \mathcal{Q}_{c_k^e}
%\end{equation}

%The model must classify each query sample using only the support examples from the same episode. This setting is referred to as an $N$-way $K$-shot classification task, where $N$ is the number of classes in the episode and $K$ is the number of labeled support examples per class.

\subsection{Prototypical Metric Learning}

Each episode is classified with a prototype-based network~\cite{snell2017prototypical} where an embedding function $f_{\phi}:\mathbb{R}^{C\times L}\!\rightarrow\!\mathbb{R}^{d}$ with learnable parameters $\phi$ maps each sample to a $d$-dimensional vector, and each class $k$ is represented by a prototype $\mathbf{c}_{k}$ computed as the mean embedding of its support examples,
\begin{equation}
    \mathbf{c}_{k}
    =
    \frac{1}{K}
    \sum_{i=1}^{K}
    f_{\phi}(\mathbf{x}^{s}_{k,i})
    \qquad k=1,\ldots,N
    \label{eq:prototype}
\end{equation}
A query $\mathbf{x}^{q}$ is assigned the class whose prototype is nearest in squared Euclidean distance, 
\begin{equation}
    d_k(\mathbf{x}^{q}) 
    = 
    \left\|f_{\phi}(\mathbf{x}^{q}) - \mathbf{c}_{k}\right\|_2^2
    \label{eq:proto_distance}
\end{equation}
with class probabilities given by a softmax over negative distances:
\begin{equation}
    p_{\phi}(y=c_k \mid \mathbf{x}^{q})
    =
    \frac{\exp(-d_k(\mathbf{x}^{q}))}
    {\sum_{j=1}^{N}\exp(-d_j(\mathbf{x}^{q}))}
    \label{eq:proto_probability}
\end{equation}
The objective of an episode is to minimize the average negative log-likelihood over query samples,
\begin{equation}
    \mathcal{L}_{\mathrm{epi}}(\phi)
    =
    -\frac{1}{|\mathcal{Q}|}
    \sum_{(\mathbf{x}^{q},y^{q})\in\mathcal{Q}}
    \log p_{\phi}(y=y^{q}\mid\mathbf{x}^{q})
    \label{eq:episodic_loss}
\end{equation}

\subsection{Omni-Scale Convolution and Receptive Fields}

Discriminative patterns in time series may appear at different temporal lengths, so the convolutional encoder should cover a broad range of scales without dataset-specific kernel-size tuning. The Omni-Scale CNN (OS-CNN)~\cite{tang2020omni} achieves this by stacking prime-sized 1D kernels whose summed receptive fields can realize essentially any temporal span.

Let $\mathcal{P}^{(j)}$ denote the kernel-size set at the $j$-th OS stage ($j=1,2,3$):
\begin{equation}
    \mathcal{P}^{(j)}
    =
    \begin{cases}
        \{p : p \text{ prime},\ p \leq p_k\}, & j=1,2 \\
        \{1,2\}, & j=3
    \end{cases}
    \label{eq:os_kernel_set}
\end{equation}
where $p_k$ is the largest prime kernel size allowed in the first two stages. For kernels $p^{(j)} \in \mathcal{P}^{(j)}$ with stride $1$ and no dilation, the receptive-field length of a single three-stage path is
\begin{equation}
    S = p^{(1)} + p^{(2)} + p^{(3)} - 2
    \label{eq:os_rf}
\end{equation}
Combining prime sizes in the first two stages with $\{1,2\}$ in the third allows the block to cover both even and adjacent odd receptive-field lengths. We use this design as the local multi-scale feature extraction branch of the OSSE-LSTM encoder.

\subsection{Channel Recalibration}
\label{subsec:se}
 
The OS-CNN branch produces multi-kernel features whose channels may not
contribute equally to the final representation. To emphasize informative channels
and suppress less useful ones, we apply a Squeeze-and-Excitation (SE)
block~\cite{hu2018squeeze}. Given a feature map
$\mathbf{H}\in\mathbb{R}^{D\times L'}$ with $D$ channels and temporal length
$L'$, the \emph{squeeze} step averages each channel over time to form a
descriptor $\mathbf{z}\in\mathbb{R}^{D}$. The \emph{excitation} step maps
$\mathbf{z}$ through a two-layer bottleneck with reduction ratio
$r_{\mathrm{se}}$, a ReLU, and a sigmoid to obtain channel weights
$\mathbf{a}\in(0,1)^{D}$. Each channel is rescaled by its weight, giving the
recalibrated map $\widetilde{\mathbf{H}}$. We denote this operation
$\mathrm{SE}(\cdot)$ in the encoder description of Section~\ref{sec:methodology}.

\subsection{Bidirectional Recurrent Temporal Modeling}
\label{subsec:bilstm}
 
The OS-CNN branch captures local multi-scale patterns through finite receptive
fields but does not maintain a recurrent state along the time axis. To complement
it, we use a Bidirectional Long Short-Term Memory (BiLSTM)
branch~\cite{hochreiter1997long,schuster1997bidirectional}. A forward LSTM reads
the input $\mathbf{x}_{:,t}\in\mathbb{R}^{C}$ from $t=1$ to $L$ while a backward
LSTM reads it in reverse, each maintaining hidden and memory states of dimension
$h$. Concatenating the two directional hidden states yields
$\mathbf{h}^{\mathrm{bi}}_{t}\in\mathbb{R}^{2h}$, which is averaged over time to
produce a fixed-length representation
$\mathbf{v}_{\mathrm{BiLSTM}}=\frac{1}{L}\sum_{t=1}^{L}\mathbf{h}^{\mathrm{bi}}_{t}
\in\mathbb{R}^{2h}$, fused with the omni-scale convolutional representation in
the OSSE-LSTM encoder.

\subsection{Integrated Gradients for Input Attribution}
\label{subsec:ig}
 
Integrated Gradients (IG)~\cite{sundararajan2017axiomatic} assigns an importance
score to each input coordinate by integrating the gradient of a scalar model
output along a straight-line path from a baseline $\mathbf{x}'\in\mathbb{R}^{C\times L}$
to the target input $\mathbf{x}$, scaling the result by
$(\mathbf{x}-\mathbf{x}')$. In standard classifiers the scalar output is the
logit or probability of a target class. For prototype-based classification,
however, the decision depends on distances between an input embedding and class
prototypes rather than on a fixed classifier logit~\cite{snell2017prototypical},
so applying IG to Prototypical Networks requires a scalar function that reflects
this distance-based rule. We define such a function in
Section~\ref{sec:methodology}.

\section{Methodology}
\label{sec:methodology}
\begin{figure*}[htbp]
    \centering
    \includegraphics[width=0.76\textwidth]{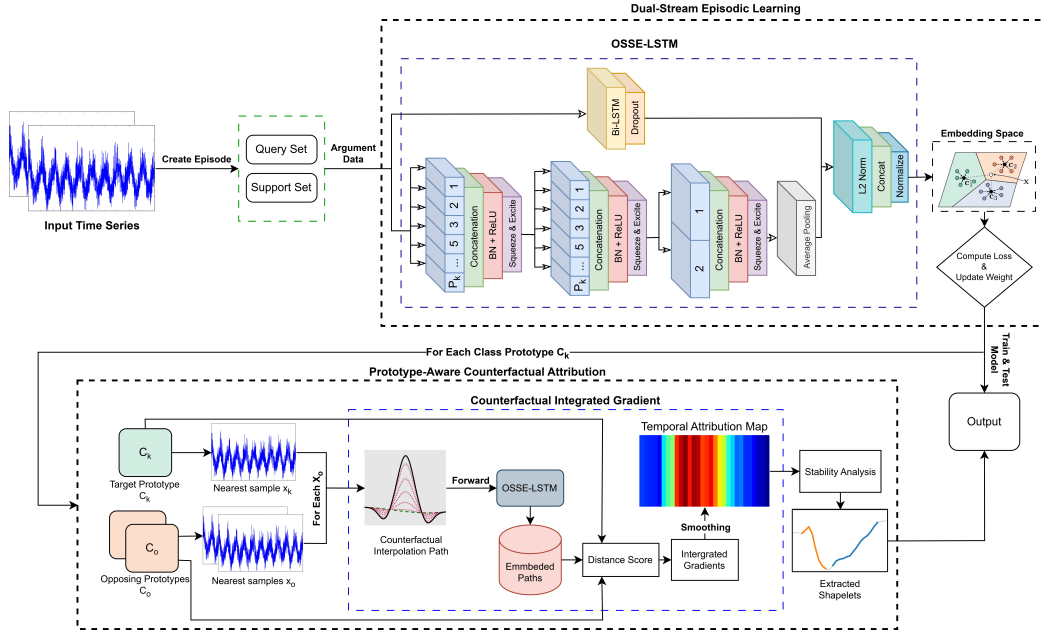}
    \caption{Overall workflow of OSSE-LSTM. %Phase~I learns a prototype-based embedding space through dual-stream episodic training with support and query samples. Phase~II interprets the learned metric space by using target and opposing prototypes to construct counterfactual interpolation paths, from which C-IG produces temporal attribution maps explaining target-opponent class separation.
    }
    \label{fig:system_architecture}
\end{figure*}

In this section, we introduce OSSE-LSTM, a framework for few-shot time series classification that combines a dual-stream encoder for multi-scale and bidirectional temporal modeling with Counterfactual Integrated Gradients (C-IG) for prototype-aware interpretation. As illustrated in Fig.~\ref{fig:system_architecture}, OSSE-LSTM follows a two-phase workflow. Phase~I learns a prototype-oriented embedding space through episodic few-shot training with a dual-stream encoder. Phase~II interprets the learned metric space using prototype-aware Counterfactual Integrated Gradients (C-IG), which identify temporal regions driving target--opponent class separation. We additionally evaluate an optional test-time prototype refinement strategy in which C-IG attribution maps guide support augmentation without updating the trained encoder.

\subsection{Phase~I: Dual-Stream Episodic Learning}

Phase~I learns an optimal embedding function $f_{\phi}:\mathbb{R}^{C\times L}\rightarrow\mathbb{R}^{d}$ for accurate prototype-based few-shot classification. The encoder has two complementary streams: an OSSE convolutional stream for multi-scale local features and a BiLSTM stream for bidirectional temporal context. Their outputs are fused into a normalized embedding and trained with the episodic objective in Eq.~\eqref{eq:episodic_loss}.

\subsubsection{Omni-Scale Squeeze-and-Excitation (OSSE) Convolutional Stream}

The OSSE stream applies a stack of $L_{\mathrm{OS}}$ layers, each performing parallel multi-kernel convolution, batch normalization, ReLU, and SE recalibration to capture features across multiple receptive-field sizes.

Let $\mathbf{H}^{(0)}=\mathbf{x}$ and let $\mathbf{H}^{(\ell-1)}\in\mathbb{R}^{D_{\ell-1}\times L_{\ell-1}}$ be the input to the $\ell$-th OSSE layer with kernel-size set $\mathcal{P}^{(\ell)}$ from Eq.~\eqref{eq:os_kernel_set}. For each kernel size $r\in\mathcal{P}^{(\ell)}$, a 1D temporal convolution $\mathrm{Conv}^{(\ell,r)}(\cdot)$ produces a feature map; outputs are then concatenated along the channel axis,
\begin{equation}
    \mathbf{U}^{(\ell)}
    =
    \mathrm{Concat}_{r\in\mathcal{P}^{(\ell)}}\!\left[\mathrm{Conv}^{(\ell,r)}\!\left(\mathbf{H}^{(\ell-1)}\right)\right]
    \label{eq:method_osse_concat}
\end{equation}
normalized, activated, and recalibrated by the SE block: %from Eqs.~\eqref{eq:se_squeeze}--\eqref{eq:se_scale}:
\begin{equation}
    \mathbf{H}^{(\ell)}
    =
    \mathrm{SE}\!\left(\rho\!\left(\mathrm{BN}\!\left(\mathbf{U}^{(\ell)}\right)\right)\right)
    \label{eq:method_osse_se}
\end{equation}
where $\rho(\cdot)$ is the ReLU and $\mathrm{BN}(\cdot)$ is batch normalization. Zero padding preserves the temporal length within each layer.

For efficient implementation, the parallel kernels in $\mathcal{P}^{(\ell)}$ are realized as a single masked convolution of size $r_{\max}^{(\ell)}=\max\mathcal{P}^{(\ell)}$: smaller kernels are centered inside $r_{\max}^{(\ell)}$ and positions outside their valid span are zeroed by a binary mask, equivalent to parallel OS kernels. %, equivalent to the parallel formulation~\cite{tang2020omni}.

After the final layer, $\mathbf{H}_{\mathrm{CNN}}=\mathbf{H}^{(L_{\mathrm{OS}})}\in\mathbb{R}^{D_c\times L_c}$ is reduced to a fixed-length vector by temporal average pooling,
\begin{equation}
    v_{\mathrm{CNN},d}
    =
    \frac{1}{L_c}\sum_{t=1}^{L_c} H_{\mathrm{CNN},d,t},
    \qquad d=1,\ldots,D_c
    \label{eq:method_cnn_vector}
\end{equation}
yielding $\mathbf{v}_{\mathrm{CNN}} = [v_{\mathrm{CNN},1},\ldots,v_{\mathrm{CNN},D_c}]^\top$.

\subsubsection{Bidirectional Recurrent Stream}

The recurrent stream reads the input time series $\mathbf{x}_{:,t}\in\mathbb{R}^{C}$ at each time step $t=1,\ldots,L$ and applies the BiLSTM %defined in Eqs.~\eqref{eq:bilstm1}--\eqref{eq:bilstm_pool} 
with dropout regularization on the recurrent outputs. The resulting representation $\mathbf{v}_{\mathrm{BiLSTM}}\in\mathbb{R}^{2h}$ summarizes past-to-future and future-to-past dependencies that are not fully captured by local convolutional receptive fields.

\subsubsection{Feature Fusion and Embedding Projection}

Because prototype classification uses Euclidean distances, each stream output is $L_2$-normalized before fusion:
\begin{equation}
    \mathbf{g}_{\mathrm{CNN}}
    =
    \frac{\mathbf{v}_{\mathrm{CNN}}}{\|\mathbf{v}_{\mathrm{CNN}}\|_2+\epsilon},
    \quad
    \mathbf{g}_{\mathrm{BiLSTM}}
    =
    \frac{\mathbf{v}_{\mathrm{BiLSTM}}}{\|\mathbf{v}_{\mathrm{BiLSTM}}\|_2+\epsilon}
    \label{eq:method_branch_norm}
\end{equation}
where $\epsilon$ ensures numerical stability. The two are concatenated as $\mathbf{g}=[\mathbf{g}_{\mathrm{CNN}}\,\|\,\mathbf{g}_{\mathrm{BiLSTM}}]$, and the final embedding is produced by layer normalization followed by $L_2$ normalization,
\begin{equation}
    f_{\phi}(\mathbf{x})
    =
    \frac{\mathrm{LN}(\mathbf{g})}{\|\mathrm{LN}(\mathbf{g})\|_2+\epsilon}
    \in\mathbb{R}^{d}
    \label{eq:method_final_embedding}
\end{equation}
with $\phi=\{\phi_c,\phi_r\}$ collecting both stream parameters. This embedding feeds the prototypes, query probabilities, and loss in Eqs.~\eqref{eq:prototype}--\eqref{eq:episodic_loss}.

\subsubsection{Episodic Optimization}

Each training episode samples $N$ classes with $K$ support and $R$ query examples per class. All are encoded by the shared $f_{\phi}$, and support embeddings form prototypes via Eq.~\eqref{eq:prototype}, query probabilities follow from Eq.~\eqref{eq:proto_probability}, and $\phi$ is updated by minimizing Eq.~\eqref{eq:episodic_loss}. 
%Algorithm~\ref{alg:episodic_training} summarizes the described procedure. 

At inference, the trained encoder is fixed: per-episode prototypes are computed from the support set and queries are classified using the same prototype-distance rule without updating model
parameters.

\iffalse
\begin{algorithm}[htbp]
\small
\caption{Episodic Optimization of OSSE-LSTM}
\label{alg:episodic_training}
\begin{algorithmic}[1]
    \Require Training dataset $\mathcal{D}_{\mathrm{train}}$, episode configuration $(N,K,R)$, encoder $f_{\phi}$
    \For{each training episode}
        \State Sample class set $\mathcal{Y}_e$ with $N$ classes
        \State Construct support set $\mathcal{S}$ ($K$ per class) and query set $\mathcal{Q}$ ($R$ per class)
        \State Compute embeddings for $\mathcal{S}\cup\mathcal{Q}$ via Eq.~\eqref{eq:method_final_embedding}
        \State Compute prototypes $\{\mathbf{c}_k\}_{k=1}^{N}$ via Eq.~\eqref{eq:prototype}
        \State Compute query probabilities via Eq.~\eqref{eq:proto_probability}
        \State Compute episodic loss $\mathcal{L}_{\mathrm{epi}}(\phi)$ via Eq.~\eqref{eq:episodic_loss}
        \State Update $\phi$ by backpropagation
    \EndFor
    \State \Return Trained encoder $f_{\phi}$
\end{algorithmic}
\end{algorithm}
\fi 

\subsection{Phase~II: Prototype-Aware Counterfactual Attribution}

After training, the encoder defines a metric space in which each class is represented by a prototype. Phase~II explains decisions in the original time domain by attributing \emph{prototype separation} rather than a classifier logit, identifying input regions that move the input $\mathbf{x}$ closer to its target prototype and farther from opposing ones. Algorithm~\ref{alg:prototype_cig} summarizes the procedure, and the following subsections explain the process. %the underlying definitions follow.

\begin{algorithm}[htbp]
\small
\caption{Prototype-Aware C-IG Attribution}
\label{alg:prototype_cig}
\begin{algorithmic}[1]
    \Require Trained encoder $f_{\phi}$, target input $\mathbf{x}$, target prototype $\mathbf{c}_k$, opposing prototypes $\{\mathbf{c}_o\}_{o\neq k}$, opposing baselines $\{\mathbf{b}_{k,o}\}_{o\neq k}$, steps $m$
    \For{each opposing class $o\neq k$}
        \State Compute prototype-margin score $G_{k,o}$ via Eq.~\eqref{eq:method_proto_margin}
        \State Construct counterfactual path from $\mathbf{b}_{k,o}$ to $\mathbf{x}$ via Eq.~\eqref{eq:method_cig_path}
        \State Sample interpolation coefficients via Eq.~\eqref{eq:method_cig_alpha}
        \State Compute pairwise attribution map $\mathbf{A}_{k,o}$ via Eq.~\eqref{eq:method_pairwise_cig}
        \State Smooth and normalize to obtain $\widehat{\mathbf{A}}_{k,o}$ via Eq.~\eqref{eq:method_pairwise_norm}
    \EndFor
    \State Compute opponent weights $\omega_{k,o}$ via Eq.~\eqref{eq:method_opponent_weight}
    \State Fuse into class-level map $\mathbf{M}_k$ via Eq.~\eqref{eq:method_fused_attr}
    \State Optionally reduce $\mathbf{M}_k$ to temporal curve $m_k(t)$ via Eq.~\eqref{eq:method_temporal_curve}
    \State \Return Class-level attribution map $\mathbf{M}_k$
\end{algorithmic}
\end{algorithm}

\subsubsection{Prototype-Margin Score}

For target class $k$ and opposing class $o$ with prototypes $\mathbf{c}_k$ and $\mathbf{c}_o$, respectively, the prototype-margin score is
\begin{equation}
    G_{k,o}(\mathbf{x}) = d_o(\mathbf{x}) - d_k(\mathbf{x})
    \label{eq:method_proto_margin}
\end{equation}
where $d_k(\mathbf{x})$ and $d_o(\mathbf{x})$ are the prototype distances from Eq.~\eqref{eq:proto_distance}. Larger $G_{k,o}$ means $\mathbf{x}$ is closer to the target than to the opponent, so attributing $G_{k,o}$ identifies regions supporting class $k$ over $o$.

\subsubsection{Counterfactual Interpolation Path}

Instead of an artificial zero baseline, our C-IG component uses an opposing-class 
reference. Let $\mathbf{b}_{k,o}\in\mathbb{R}^{C\times L}$ be a representative input from the opposing class $o$. The interpolation 
path from this reference to the target input $\mathbf{x}$ is
\begin{equation}
    \mathbf{x}_{k,o}(\alpha)
    =
    \mathbf{b}_{k,o} + \alpha(\mathbf{x}-\mathbf{b}_{k,o}), 
    \qquad \alpha\in[0,1]
    \label{eq:method_cig_path}
\end{equation}
gradually transforming an opposing-class signal into the target. 
To approximate the path integral, we uniformly sample $m$ interpolation 
coefficients,
\begin{equation}
    \alpha_i = \frac{i}{m-1}, 
    \qquad i=0,\ldots,m-1
    \label{eq:method_cig_alpha}
\end{equation}
following the standard IG discretization.

\subsubsection{Pairwise C-IG Attribution}

For each target--opponent pair, the pairwise attribution map averages prototype-margin gradients along the counterfactual path,
\begin{equation}
    \mathbf{A}_{k,o}
    =
    \left|
    (\mathbf{x}-\mathbf{b}_{k,o})
    \odot
    \frac{1}{m}\sum_{i=0}^{m-1} \nabla_{\mathbf{x}} G_{k,o}\!\left(\mathbf{x}_{k,o}(\alpha_i)\right)
    \right|
    \label{eq:method_pairwise_cig}
\end{equation}
where $\odot$ is element-wise multiplication. Averaging gradients over multiple path samples suppresses dependence on a single local gradient and highlights regions that consistently increase target--opponent separation. The raw map is smoothed temporally and normalized to $[0,1]$,
\begin{equation}
    \widehat{\mathbf{A}}_{k,o}
    =
    \mathrm{Norm}_{[0,1]}\!\left(\mathbf{A}_{k,o} \circledast \mathbf{K}_{w}\right)
    \label{eq:method_pairwise_norm}
\end{equation}
where $\mathbf{K}_{w}$ is a uniform moving-average kernel of width $w$ and $\circledast$ denotes temporal convolution. Smoothing suppresses isolated spikes, and normalization makes pairwise maps comparable between opponents.

\subsubsection{Multi-Opponent Fusion}

In an $N$-way episode, a target class $k$ must be separated from multiple opponents. The C-IG computes one pairwise map per opponent class $o$ ($o\neq k$), and fuses them into a class-level map, weighting each opponent by its proximity to the target prototype:
\begin{equation}
    \omega_{k,o}
    =
    \frac{\exp(-\|\mathbf{c}_k-\mathbf{c}_o\|_2)}{\sum_{j\neq k}\exp(-\|\mathbf{c}_k-\mathbf{c}_j\|_2)}
    \label{eq:method_opponent_weight}
\end{equation}
Closer opponents receive larger weights because they represent harder class boundaries. The fused map is
\begin{equation}
    \mathbf{M}_{k}
    =
    \mathrm{Norm}_{[0,1]}\!\left(\sum_{o\neq k} \omega_{k,o}\,\widehat{\mathbf{A}}_{k,o}\right) \in [0,1]^{C\times L}
    \label{eq:method_fused_attr}
\end{equation}
which highlights regions that separate target class $k$ from its most relevant opponents. For visualization, the multivariate map is reduced to a temporal importance curve,
\begin{equation}
    m_k(t) = \max_{1\leq c\leq C} \mathbf{M}_{k}(c,t)
    \label{eq:method_temporal_curve}
\end{equation}
whose highlighted intervals are interpreted as temporal evidence derived from attribution, not as conventional shapelets learned by a separate shapelet-mining algorithm.

\subsection{Attribution-Guided Test-Time Prototype Refinement}

C-IG attribution can additionally serve as a soft mask for test-time support augmentation. This refinement is used only at inference, and does not update the trained encoder. Its purpose is to reduce the sensitivity of a few-shot prototype to irrelevant background variation by preserving high-attribution regions and perturbing low-attribution ones. %Algorithm~\ref{alg:cig_tta} summarizes the procedure.

For each support sample $\mathbf{x}_i\in\mathcal{S}_{k}$, an episode-local opposing baseline is obtained by averaging the support samples of all other classes,
\begin{equation}
    \mathbf{b}_i
    =
    \frac{1}{|\mathcal{S}\setminus\mathcal{S}_{k}|}
    \sum_{\substack{(\mathbf{x}_j,y_j)\in\mathcal{S}\\y_j\neq k}}
    \mathbf{x}_j
    \label{eq:method_tta_baseline}
\end{equation}
To measure separation against all opponents jointly, we define the multi-opponent margin
\begin{equation}
    G_{k}^{\mathrm{TTA}}(\mathbf{x})
    =
    \frac{1}{N-1}\sum_{o\neq k} d_o(\mathbf{x}) - d_k(\mathbf{x})
    \label{eq:method_tta_margin}
\end{equation}
which increases as $\mathbf{x}$ moves farther from opposing prototypes and closer to its own. The interpolation path and its discretization are computed as
\begin{equation}
    \mathbf{x}_{i}^{(r)} = \mathbf{b}_i + \alpha_r(\mathbf{x}_i - \mathbf{b}_i), \quad \alpha_r = \frac{r}{m-1}, \quad r=0,\ldots,m-1
    \label{eq:method_tta_path}
\end{equation}
and the support-specific attribution map is
\begin{equation}
    \mathbf{A}_i^{\mathrm{TTA}}
    =
    \left|
    (\mathbf{x}_i - \mathbf{b}_i) \odot \frac{1}{m}\sum_{r=0}^{m-1} \nabla_{\mathbf{x}} G_k^{\mathrm{TTA}}\!\left(\mathbf{x}_{i}^{(r)}\right)
    \right| \in \mathbb{R}^{C\times L}
    \label{eq:method_tta_attr}
\end{equation}

To form a temporal mask, we average $\mathbf{A}_i^{\mathrm{TTA}}$ across channels,
\begin{equation}
    a_i(t) = \frac{1}{C}\sum_{c=1}^{C} A_{i,c,t}^{\mathrm{TTA}}, \qquad t=1,\ldots,L
    \label{eq:method_tta_channel_mean}
\end{equation}
min--max normalize over time,
\begin{equation}
    m_i(t)
    =
    \frac{a_i(t) - \min_t a_i(t)}{\max_t a_i(t) - \min_t a_i(t) + \epsilon}
    \label{eq:method_tta_temporal_mask}
\end{equation}
and broadcast across channels to obtain the soft mask:
\begin{equation}
    M_{i,c,t}^{\mathrm{TTA}} = m_i(t), \qquad c=1,\ldots,C
    \label{eq:method_tta_mask}
\end{equation}

The mask $\mathbf{M}^{\mathrm{TTA}}_i$ thus marks high-attribution 
regions for preservation and low-attribution regions for perturbation. 
We generate $A$ augmented support samples by injecting noise mainly 
into the latter,
\begin{equation}
    \widetilde{\mathbf{x}}^{(a)}_i 
    = 
    \mathbf{x}_i + (\mathbf{1} - \mathbf{M}^{\mathrm{TTA}}_i) \odot \boldsymbol{\eta}^{(a)}_i,
    \qquad a = 1, \ldots, A,
    \label{eq:method_tta_aug}
\end{equation}
where the noise $\boldsymbol{\eta}^{(a)}_i$ is scaled by the per-channel temporal standard deviation $\boldsymbol{\sigma}_{i}\in\mathbb{R}^{C\times 1}$ of $\mathbf{x}_i$, 
\begin{equation}
    \boldsymbol{\eta}^{(a)}_i = \beta\,\boldsymbol{\sigma}_i \odot \boldsymbol{\epsilon}^{(a)}_i,
    \qquad \boldsymbol{\epsilon}^{(a)}_i \sim \mathcal{N}(\mathbf{0}, \mathbf{I}),
    \label{eq:method_tta_noise}
\end{equation}
Here, $\beta$ controls perturbation strength. %High-attribution regions are largely preserved while low-attribution regions receive stronger perturbation.
Next, for each class $k$, the augmented support set is
\begin{equation}
    \widetilde{\mathcal{S}}_{k} = \mathcal{S}_{k} \cup \{\widetilde{\mathbf{x}}_{i}^{(a)} \mid \mathbf{x}_i\in\mathcal{S}_{k},\, a=1,\ldots,A\}
    \label{eq:method_augmented_support}
\end{equation}
and the refined prototype is the mean embedding over original and augmented samples,
\begin{equation}
    \widehat{\mathbf{c}}_k = \frac{1}{|\widetilde{\mathcal{S}}_{k}|} \sum_{\mathbf{u}\in\widetilde{\mathcal{S}}_{k}} f_{\phi}(\mathbf{u})
    \label{eq:method_refined_proto}
\end{equation}

During refined inference, query probabilities are computed via Eq.~\eqref{eq:proto_probability} with $\widehat{\mathbf{c}}_k$ in place of $\mathbf{c}_k$. Unlike episodic training, this procedure does not update encoder parameters. Instead, it only recomputes the support-derived prototypes using attribution-guided augmentations.

\section{Experiments}
\label{sec:experiments}

We evaluate our OSSE-LSTM from four perspectives: classification performance under limited training data, component-level ablation, prototype-aware interpretability, and C-IG-guided test-time prototype refinement.

\subsection{Experimental Setup}

\subsubsection{Datasets}

We use $19$ univariate datasets from the UCR archive~\cite{dau2019ucr}. %covering ECG signals, sensor traces, synthetic control patterns, object outlines, and motion sequences. %and 9 multivariate datasets from the UEA archive~\cite{bagnall2018uea} in which each sample contains multiple synchronized channels. 
A univariate sample is $\mathbf{x}\in\mathbb{R}^{1\times L}$ where L is the length of the sample.
%and a multivariate sample is $\mathbf{x}\in\mathbb{R}^{C\times L}$. 
When necessary, sequences are padded to consistent shapes within episodic batches. No handcrafted features or domain-specific preprocessing are applied.

\subsubsection{Baselines} 

We compare OSSE-LSTM against six representative baselines spanning complementary
modeling paradigms. \textbf{LSTM-FCN}~\cite{karim2017lstm} pairs a recurrent
branch with a fully convolutional one, isolating the benefit of our multi-scale
OSSE encoder and prototype-based learning. \textbf{MiniRocket}~\cite{dempster2021minirocket}
represents efficient convolutional-transform methods and provides a strong
non-deep reference under limited training data. \textbf{TapNet}~\cite{zhang2020tapnet}
combines attentional prototype learning with recurrent and convolutional feature
extraction, making it the closest metric-learning baseline to ours.
\textbf{DPSN}~\cite{tang2020interpretable} learns shapelets as class prototypes
and classifies by similarity to a small set of discriminative subsequences. It is
the most directly comparable baseline, being the only competitor that targets both
label efficiency and interpretability. %and it tests whether shapelet evidence alone suffices or whether a dual-stream embedding with prototype-margin attribution yields stronger prototypes. 
\textbf{TS2Vec}~\cite{yue2022ts2vec} learns
self-supervised contextual representations at multiple temporal scales, testing
whether episodic training offers an advantage over general-purpose representation
learning. Finally, \textbf{InceptionTime}~\cite{ismail2020inceptiontime} is a
strong multi-scale convolutional architecture that isolates the contribution of
recurrent temporal modeling beyond convolutional multi-scale extraction alone.
%Together, these baselines span hybrid CNN-RNN modeling, convolutional transforms, prototype-based metric learning, shapelet-based few-shot classification, self-supervised representation learning, and deep multi-scale convolution.

\subsubsection{Evaluation Protocol}

All experiments follow the episodic protocol introduced in Section~\ref{sec:preliminaries}. For each episode, support samples are used to construct class prototypes in the learned embedding space, while query samples are classified according to their distances to these prototypes. To examine robustness under different levels of training-data availability, we vary the training examples $K \in\{2, 4, 6, 8\}$. Results are averaged over repeated runs of test episodes and are reported as Top-1 classification accuracy (\%) $\pm$ standard deviation.

\subsubsection{Implementation Details}

The OSSE branch uses a maximum receptive-field range of $\min(\lfloor L/4 \rfloor, 89)$, where $L$ is the input sequence length, and $89$ is the largest prime considered, limiting kernel size on long sequences while preserving multi-scale coverage. The BiLSTM branch uses 16 hidden units per direction (32-dimensional output) with dropout $0.5$ applied during training. Models are optimized with Adam (learning rate $10^{-3}$, weight decay $10^{-5}$). All evaluations use prototype-based inference without update of test-time parameters.

\subsection{Effect of Support Set Size on Classification Results}

%\subsubsection{Effect of Support Set Size}

\begin{table*}[t]
\centering
\caption{Comparison of classification accuracy (\%) $\pm$ STD under 2-shot and 4-shot settings at full training ratio.}
\label{tab:shot_2_4}
\setlength{\tabcolsep}{3pt}
\renewcommand{\arraystretch}{1.05}
\resizebox{\textwidth}{!}{%
\begin{tabular}{lcccccccccccccc}
\toprule
\textbf{Dataset} & \multicolumn{7}{c}{\textbf{2-Shot}} & \multicolumn{7}{c}{\textbf{4-Shot}} \\
\cmidrule(lr){2-8} \cmidrule(lr){9-15}
 & \textbf{InceptionTime} & \textbf{LSTM-FCN} & \textbf{MiniRocket} & \textbf{TapNet} & \textbf{TS2Vec} & \textbf{DPSN} & \textbf{OSSE-LSTM}
 & \textbf{InceptionTime} & \textbf{LSTM-FCN} & \textbf{MiniRocket} & \textbf{TapNet} & \textbf{TS2Vec} & \textbf{DPSN} & \textbf{OSSE-LSTM} \\
\midrule

ArrowHead
& 78.53$\pm$3.06 & 76.01$\pm$1.57 & 76.14$\pm$5.04 & \textbf{83.33$\pm$1.35} & 58.77$\pm$1.04 & 76.58$\pm$1.04 & 81.95$\pm$3.08
& 78.30$\pm$3.44 & 70.87$\pm$3.06 & 76.50$\pm$3.73 & \textbf{86.00$\pm$1.27} & 64.95$\pm$1.04 & 79.40$\pm$0.77 & 83.82$\pm$2.32 \\

BME
& 99.92$\pm$0.10 & 99.97$\pm$0.08 & 94.13$\pm$2.21 & 99.60$\pm$0.26 & 53.06$\pm$1.67 & 83.58$\pm$1.27 & \textbf{100.00$\pm$0.00}
& 99.63$\pm$0.12 & 99.68$\pm$0.31 & 96.03$\pm$3.16 & 99.46$\pm$0.25 & 58.88$\pm$2.97 & 84.61$\pm$0.87 & \textbf{100.00$\pm$0.00} \\

CBF
& 99.11$\pm$0.68 & 99.29$\pm$0.11 & 96.84$\pm$0.99 & 99.16$\pm$0.25 & 93.65$\pm$0.52 & 97.19$\pm$0.31 & \textbf{99.97$\pm$0.04}
& 98.90$\pm$0.51 & 99.25$\pm$0.19 & 96.13$\pm$1.37 & 98.89$\pm$0.19 & 96.12$\pm$0.47 & 98.08$\pm$0.26 & \textbf{99.95$\pm$0.06} \\

Chinatown
& 96.00$\pm$1.37 & 98.25$\pm$0.28 & 95.02$\pm$1.85 & 97.49$\pm$0.82 & 89.46$\pm$1.48 & 80.51$\pm$0.86 & \textbf{98.65$\pm$0.38}
& 75.30$\pm$5.21 & 98.20$\pm$0.35 & 93.93$\pm$1.25 & 97.03$\pm$0.64 & 95.12$\pm$0.84 & 85.10$\pm$0.71 & \textbf{98.81$\pm$0.25} \\

ECG200
& 89.30$\pm$1.10 & 87.04$\pm$0.83 & 87.30$\pm$1.39 & 88.57$\pm$1.31 & 62.95$\pm$0.98 & 87.49$\pm$0.67 & \textbf{89.97$\pm$1.57}
& 79.03$\pm$3.10 & 87.00$\pm$0.86 & 86.73$\pm$3.43 & \textbf{90.14$\pm$0.81} & 67.30$\pm$1.40 & 89.24$\pm$0.62 & 89.97$\pm$1.93 \\

GunPointAgeSpan
& 98.89$\pm$0.39 & 98.93$\pm$0.71 & 96.85$\pm$0.26 & 97.04$\pm$0.46 & 64.70$\pm$1.18 & 98.79$\pm$0.26 & \textbf{99.77$\pm$0.21}
& 99.36$\pm$0.20 & 99.15$\pm$0.54 & 96.11$\pm$0.93 & 96.94$\pm$0.41 & 74.07$\pm$1.25 & 99.27$\pm$0.13 & \textbf{99.73$\pm$0.26} \\

GunPointMaleVersusFemale
& 99.20$\pm$0.45 & 96.22$\pm$1.52 & 99.19$\pm$0.70 & 99.71$\pm$0.18 & 78.71$\pm$1.84 & 96.08$\pm$0.34 & \textbf{100.00$\pm$0.00}
& 99.81$\pm$0.08 & 99.66$\pm$0.33 & 98.54$\pm$0.97 & 99.86$\pm$0.18 & 89.31$\pm$1.63 & 96.75$\pm$0.26 & \textbf{100.00$\pm$0.00} \\

GunPointOldVersusYoung
& 98.38$\pm$1.49 & 99.99$\pm$0.02 & 99.99$\pm$0.02 & 99.94$\pm$0.12 & 93.15$\pm$0.65 & 97.38$\pm$0.55 & \textbf{100.00$\pm$0.00}
& 99.85$\pm$0.30 & \textbf{100.00$\pm$0.00} & 99.98$\pm$0.03 & \textbf{100.00$\pm$0.00} & 97.97$\pm$0.52 & 96.97$\pm$0.46 & \textbf{100.00$\pm$0.00} \\

ItalyPowerDemand
& 96.56$\pm$0.45 & 96.47$\pm$0.41 & 94.97$\pm$0.58 & \textbf{96.66$\pm$0.31} & 80.54$\pm$1.56 & 89.18$\pm$0.58 & 94.98$\pm$0.58
& \textbf{96.45$\pm$0.32} & 95.85$\pm$0.48 & 93.42$\pm$1.25 & 95.90$\pm$0.38 & 86.27$\pm$0.97 & 89.89$\pm$0.80 & 95.15$\pm$0.76 \\

MoteStrain
& 79.81$\pm$3.45 & \textbf{94.28$\pm$0.62} & 90.50$\pm$0.94 & 84.59$\pm$0.55 & 79.68$\pm$2.22 & 81.43$\pm$0.65 & 90.69$\pm$1.82
& 75.26$\pm$3.43 & \textbf{93.67$\pm$0.71} & 89.73$\pm$0.89 & 85.24$\pm$1.29 & 85.81$\pm$1.00 & 84.07$\pm$0.67 & 90.46$\pm$0.74 \\

Plane
& 86.64$\pm$3.85 & 99.99$\pm$0.02 & 98.38$\pm$0.31 & 99.28$\pm$0.56 & 98.70$\pm$0.19 & 99.35$\pm$0.17 & \textbf{100.00$\pm$0.00}
& 85.85$\pm$3.90 & 99.95$\pm$0.14 & 97.78$\pm$0.66 & 99.26$\pm$0.64 & 99.13$\pm$0.19 & 99.78$\pm$0.13 & \textbf{100.00$\pm$0.00} \\

SmoothSubspace
& 96.93$\pm$0.78 & 99.11$\pm$0.58 & 79.16$\pm$1.72 & 96.39$\pm$1.18 & 69.95$\pm$1.72 & 44.46$\pm$0.64 & \textbf{99.96$\pm$0.14}
& 96.21$\pm$1.12 & 98.77$\pm$0.59 & 80.08$\pm$3.85 & 96.31$\pm$1.07 & 78.45$\pm$4.34 & 48.53$\pm$0.85 & \textbf{99.91$\pm$0.28} \\

SonyAIBORobotSurface1
& 84.07$\pm$1.94 & 95.16$\pm$0.86 & 91.46$\pm$1.58 & 92.74$\pm$1.51 & 86.72$\pm$1.28 & 53.00$\pm$0.28 & \textbf{98.13$\pm$0.62}
& 86.52$\pm$2.09 & 95.88$\pm$0.65 & 89.28$\pm$1.86 & 91.53$\pm$1.94 & 91.45$\pm$0.96 & 54.91$\pm$0.49 & \textbf{97.72$\pm$0.57} \\

SonyAIBORobotSurface2
& 86.11$\pm$2.30 & 96.27$\pm$0.36 & 79.71$\pm$2.94 & 84.35$\pm$1.03 & 78.17$\pm$1.96 & 83.55$\pm$0.71 & \textbf{97.69$\pm$0.59}
& 85.53$\pm$2.15 & 96.12$\pm$0.55 & 77.51$\pm$3.33 & 85.24$\pm$1.16 & 83.53$\pm$0.97 & 86.94$\pm$0.75 & \textbf{97.71$\pm$0.37} \\

SyntheticControl
& 97.92$\pm$1.02 & 98.92$\pm$0.37 & 96.58$\pm$0.60 & 99.62$\pm$0.22 & 96.36$\pm$0.76 & 90.29$\pm$0.56 & \textbf{99.92$\pm$0.15}
& 96.29$\pm$0.68 & 98.68$\pm$0.25 & 93.20$\pm$1.33 & 99.51$\pm$0.08 & 97.89$\pm$0.25 & 90.79$\pm$0.35 & \textbf{99.88$\pm$0.19} \\

ToeSegmentation1
& 88.59$\pm$3.43 & 95.22$\pm$0.85 & 80.89$\pm$4.68 & 56.38$\pm$1.76 & 77.73$\pm$1.36 & 94.56$\pm$0.31 & \textbf{96.64$\pm$0.74}
& 83.78$\pm$6.03 & 95.44$\pm$0.74 & 76.71$\pm$2.57 & 61.61$\pm$1.06 & 81.96$\pm$1.87 & 95.09$\pm$0.28 & \textbf{97.03$\pm$0.95} \\

TwoLeadECG
& 77.59$\pm$7.25 & 98.72$\pm$0.93 & 88.02$\pm$3.10 & 99.07$\pm$0.46 & 75.15$\pm$1.41 & 94.20$\pm$0.78 & \textbf{99.96$\pm$0.05}
& 75.95$\pm$9.64 & 96.74$\pm$2.87 & 88.89$\pm$4.91 & 97.02$\pm$1.78 & 80.03$\pm$1.09 & 95.31$\pm$0.51 & \textbf{99.92$\pm$0.11} \\

UMD
& 98.47$\pm$0.41 & 99.12$\pm$0.18 & 91.92$\pm$2.11 & 98.82$\pm$0.60 & 64.87$\pm$1.09 & 91.08$\pm$0.79 & \textbf{99.28$\pm$0.16}
& 95.83$\pm$2.32 & 99.14$\pm$0.29 & 88.66$\pm$2.00 & 98.46$\pm$0.57 & 67.60$\pm$1.09 & 91.51$\pm$0.41 & \textbf{99.28$\pm$0.12} \\

Wine
& 62.85$\pm$5.82 & 56.84$\pm$2.42 & 50.03$\pm$0.04 & \textbf{90.70$\pm$1.21} & 55.35$\pm$1.87 & 64.74$\pm$2.20 & 83.22$\pm$3.69
& 58.69$\pm$6.03 & 55.35$\pm$2.01 & 50.03$\pm$0.04 & \textbf{91.17$\pm$0.52} & 59.12$\pm$0.00 & 66.29$\pm$1.17 & 84.93$\pm$4.12 \\

\midrule
\textbf{Average Accuracy}
& 90.26 & 93.99 & 88.79 & 92.81 & 76.72 & 84.39 & \textbf{96.36}
& 87.71 & 93.65 & 87.85 & 93.13 & 81.84 & 85.92 & \textbf{96.54} \\

\textbf{Win Count}
& 0 & 1 & 0 & 3 & 0 & 0 & \textbf{15}
& 1 & 2 & 0 & 4 & 0 & 0 & \textbf{14} \\

\bottomrule
\end{tabular}%
}
%\vspace{0.1in}
%\footnotesize{Bold indicates the best result for each dataset and shot setting at the displayed two-decimal precision; ties are bolded.}
\end{table*}

\begin{table*}[t]
\centering
\caption{Comparison of classification accuracy (\%) $\pm$ STD under 6-shot and 8-shot settings at full training ratio.}
\label{tab:shot_6_8}
\setlength{\tabcolsep}{3pt}
\renewcommand{\arraystretch}{1.05}
\resizebox{\textwidth}{!}{%
\begin{tabular}{lcccccccccccccc}
\toprule
\textbf{Dataset} & \multicolumn{7}{c}{\textbf{6-Shot}} & \multicolumn{7}{c}{\textbf{8-Shot}} \\
\cmidrule(lr){2-8} \cmidrule(lr){9-15}
 & \textbf{InceptionTime} & \textbf{LSTM-FCN} & \textbf{MiniRocket} & \textbf{TapNet} & \textbf{TS2Vec} & \textbf{DPSN} & \textbf{OSSE-LSTM}
 & \textbf{InceptionTime} & \textbf{LSTM-FCN} & \textbf{MiniRocket} & \textbf{TapNet} & \textbf{TS2Vec} & \textbf{DPSN} & \textbf{OSSE-LSTM} \\
\midrule

ArrowHead
& 73.83$\pm$5.50 & 66.51$\pm$4.13 & 77.55$\pm$4.37 & 84.53$\pm$1.78 & 66.84$\pm$1.31 & 79.91$\pm$1.10 & \textbf{85.45$\pm$2.66}
& 56.18$\pm$5.40 & 62.58$\pm$5.93 & 75.31$\pm$4.16 & \textbf{85.60$\pm$1.20} & 69.29$\pm$0.68 & 80.75$\pm$0.69 & 84.08$\pm$2.47 \\

BME
& 93.02$\pm$7.15 & 99.19$\pm$0.57 & 95.84$\pm$2.00 & 99.73$\pm$0.22 & 66.71$\pm$4.32 & 86.89$\pm$1.79 & \textbf{100.00$\pm$0.00}
& 90.83$\pm$7.47 & 98.92$\pm$0.58 & 96.29$\pm$2.96 & 99.76$\pm$0.25 & 70.79$\pm$7.49 & 87.27$\pm$2.03 & \textbf{100.00$\pm$0.00} \\

CBF
& 82.30$\pm$2.87 & 99.28$\pm$0.19 & 95.29$\pm$2.28 & 98.75$\pm$0.39 & 96.82$\pm$0.30 & 98.16$\pm$0.22 & \textbf{99.91$\pm$0.13}
& 33.47$\pm$0.34 & 99.27$\pm$0.24 & 38.91$\pm$6.99 & 98.85$\pm$0.33 & 97.67$\pm$0.44 & 98.16$\pm$0.15 & \textbf{99.72$\pm$0.28} \\

Chinatown
& 71.66$\pm$1.96 & 98.39$\pm$0.36 & 93.58$\pm$1.38 & 97.96$\pm$0.40 & 96.83$\pm$0.54 & 84.24$\pm$0.56 & \textbf{98.63$\pm$0.48}
& 71.43$\pm$7.69 & 98.29$\pm$0.44 & 93.37$\pm$0.90 & 97.84$\pm$0.61 & 97.00$\pm$0.22 & 85.97$\pm$0.60 & \textbf{98.96$\pm$0.41} \\

ECG200
& 75.44$\pm$3.29 & 86.50$\pm$0.98 & 85.83$\pm$2.32 & 88.92$\pm$0.67 & 70.45$\pm$0.92 & \textbf{90.77$\pm$0.67} & 89.48$\pm$1.60
& 74.30$\pm$5.20 & 86.05$\pm$1.61 & 86.39$\pm$1.59 & \textbf{89.89$\pm$0.45} & 71.22$\pm$1.62 & 88.78$\pm$0.62 & 89.75$\pm$1.74 \\

GunPointAgeSpan
& 99.15$\pm$0.46 & 98.54$\pm$0.98 & 95.13$\pm$0.93 & 95.89$\pm$1.45 & 79.95$\pm$2.13 & 99.48$\pm$0.22 & \textbf{99.85$\pm$0.20}
& 99.04$\pm$0.74 & 98.90$\pm$1.12 & 95.38$\pm$0.78 & 97.08$\pm$0.39 & 83.40$\pm$1.51 & 99.49$\pm$0.13 & \textbf{99.70$\pm$0.35} \\

GunPointMaleVersusFemale
& 99.85$\pm$0.15 & 99.83$\pm$0.11 & 98.48$\pm$0.81 & 99.93$\pm$0.07 & 93.62$\pm$0.76 & 96.81$\pm$0.27 & \textbf{100.00$\pm$0.00}
& 99.88$\pm$0.06 & 99.77$\pm$0.10 & 98.65$\pm$0.81 & 99.91$\pm$0.04 & 95.55$\pm$0.73 & 97.26$\pm$0.52 & \textbf{100.00$\pm$0.00} \\

GunPointOldVersusYoung
& 99.50$\pm$0.79 & \textbf{100.00$\pm$0.00} & 99.97$\pm$0.05 & 99.99$\pm$0.01 & 99.54$\pm$0.12 & 97.22$\pm$0.56 & \textbf{100.00$\pm$0.00}
& 99.81$\pm$0.29 & \textbf{100.00$\pm$0.00} & \textbf{100.00$\pm$0.01} & 99.99$\pm$0.02 & 99.83$\pm$0.07 & 97.55$\pm$0.60 & \textbf{100.00$\pm$0.00} \\

ItalyPowerDemand
& 95.43$\pm$1.34 & \textbf{95.90$\pm$0.50} & 92.87$\pm$1.83 & 95.88$\pm$0.65 & 88.25$\pm$1.01 & 90.74$\pm$0.39 & 95.34$\pm$0.40
& 94.78$\pm$0.72 & \textbf{95.86$\pm$0.55} & 91.92$\pm$1.92 & 95.29$\pm$0.57 & 88.95$\pm$0.93 & 91.11$\pm$0.21 & 95.41$\pm$0.44 \\

MoteStrain
& 73.57$\pm$3.59 & \textbf{94.10$\pm$0.78} & 91.13$\pm$1.50 & 84.54$\pm$1.08 & 87.31$\pm$0.50 & 84.20$\pm$0.46 & 90.82$\pm$1.47
& 70.25$\pm$1.65 & \textbf{94.13$\pm$0.64} & 90.69$\pm$1.92 & 83.57$\pm$1.08 & 88.79$\pm$0.88 & 84.43$\pm$0.65 & 91.12$\pm$1.25 \\

Plane
& 86.22$\pm$5.47 & \textbf{100.00$\pm$0.01} & 98.04$\pm$0.74 & 99.65$\pm$0.31 & 99.40$\pm$0.26 & 99.77$\pm$0.16 & \textbf{100.00$\pm$0.00}
& 33.28$\pm$29.10 & 99.97$\pm$0.06 & 27.05$\pm$9.46 & 99.64$\pm$0.32 & 99.53$\pm$0.28 & 99.73$\pm$0.18 & \textbf{100.00$\pm$0.00} \\

SmoothSubspace
& 94.07$\pm$2.07 & 98.33$\pm$1.08 & 79.51$\pm$2.31 & 96.75$\pm$0.55 & 81.76$\pm$3.89 & 50.10$\pm$1.03 & \textbf{99.93$\pm$0.23}
& 92.96$\pm$2.78 & 97.63$\pm$1.25 & 80.76$\pm$2.83 & 96.17$\pm$0.38 & 83.47$\pm$3.32 & 50.92$\pm$1.04 & \textbf{99.95$\pm$0.17} \\

SonyAIBORobotSurface1
& 87.58$\pm$1.78 & 97.18$\pm$0.58 & 50.06$\pm$0.19 & 92.10$\pm$1.09 & 93.49$\pm$0.69 & 54.87$\pm$0.37 & \textbf{97.93$\pm$0.70}
& 87.75$\pm$2.40 & 97.35$\pm$0.61 & 50.01$\pm$0.03 & 93.30$\pm$0.52 & 94.62$\pm$0.30 & 54.94$\pm$0.39 & \textbf{98.53$\pm$0.37} \\

SonyAIBORobotSurface2
& 83.90$\pm$1.79 & 95.28$\pm$1.03 & 77.60$\pm$3.30 & 85.54$\pm$1.21 & 86.71$\pm$1.54 & 86.74$\pm$0.72 & \textbf{97.91$\pm$0.54}
& 83.61$\pm$1.73 & 96.22$\pm$0.57 & 77.36$\pm$2.14 & 84.17$\pm$1.60 & 86.91$\pm$0.72 & 87.15$\pm$0.61 & \textbf{97.75$\pm$0.63} \\

SyntheticControl
& 93.46$\pm$1.49 & 98.57$\pm$0.29 & 93.65$\pm$1.64 & 99.46$\pm$0.15 & 98.60$\pm$0.14 & 91.59$\pm$0.24 & \textbf{99.84$\pm$0.27}
& 91.07$\pm$2.25 & 98.62$\pm$0.20 & 92.02$\pm$1.20 & 99.69$\pm$0.13 & 98.97$\pm$0.20 & 91.68$\pm$0.45 & \textbf{99.94$\pm$0.13} \\

ToeSegmentation1
& 91.00$\pm$2.33 & 94.69$\pm$0.85 & 76.70$\pm$2.52 & 61.70$\pm$1.25 & 85.47$\pm$2.69 & 95.10$\pm$0.44 & \textbf{97.11$\pm$0.60}
& 86.93$\pm$2.79 & 94.58$\pm$0.60 & 74.78$\pm$1.75 & 60.26$\pm$0.65 & 86.28$\pm$0.94 & 95.74$\pm$0.43 & \textbf{96.93$\pm$0.59} \\

TwoLeadECG
& 73.03$\pm$5.99 & 95.25$\pm$5.37 & 87.03$\pm$4.92 & 97.26$\pm$0.64 & 81.69$\pm$1.23 & 95.43$\pm$0.15 & \textbf{99.93$\pm$0.06}
& 72.51$\pm$6.10 & 96.63$\pm$3.22 & 86.89$\pm$2.28 & 90.57$\pm$1.66 & 82.43$\pm$1.33 & 94.87$\pm$1.29 & \textbf{99.91$\pm$0.10} \\

UMD
& 92.16$\pm$5.41 & 98.67$\pm$0.64 & 90.85$\pm$2.53 & 98.18$\pm$0.56 & 70.30$\pm$1.34 & 92.35$\pm$0.42 & \textbf{99.32$\pm$0.06}
& 89.47$\pm$8.77 & 98.94$\pm$0.32 & 92.93$\pm$3.57 & 98.32$\pm$0.53 & 72.99$\pm$1.58 & 91.85$\pm$0.58 & \textbf{99.34$\pm$0.06} \\

Wine
& 58.69$\pm$4.03 & 58.39$\pm$3.43 & 50.07$\pm$0.04 & \textbf{90.68$\pm$1.19} & 60.13$\pm$1.20 & 67.27$\pm$0.96 & 86.31$\pm$4.32
& 56.70$\pm$4.89 & 53.46$\pm$0.00 & 50.04$\pm$0.05 & 85.37$\pm$3.45 & 61.14$\pm$0.96 & 68.65$\pm$2.27 & \textbf{86.12$\pm$2.79} \\

\midrule
\textbf{Average Accuracy}
& 85.47 & 93.40 & 85.75 & 93.02 & 84.41 & 86.40 & \textbf{96.72}
& 78.12 & 93.01 & 78.88 & 92.38 & 85.73 & 86.65 & \textbf{96.70} \\

\textbf{Win Count}
& 0 & 4 & 0 & 1 & 0 & 1 & \textbf{15}
& 0 & 3 & 1 & 2 & 0 & 0 & \textbf{15} \\

\bottomrule
\end{tabular}%
}
%\vspace{3mm}
%\footnotesize{Bold indicates the best result for each dataset and shot setting at the displayed two-decimal precision; ties are bolded.}
\end{table*}

Tables \ref{tab:shot_2_4} and \ref{tab:shot_6_8} report classification performance as the support size varies from 2 to 8 labeled examples per class at the full training ratio. %Table I presents the 2-shot and 4-shot results, while Table II presents the 6-shot and 8-shot results. 
Across these settings, OSSE-LSTM achieves the highest average accuracy at every support size, with 96.36\%, 96.54\%, 96.72\%, and 96.70\% for 2-, 4-, 6- and 8-shot, respectively. It also obtains the highest per-dataset win count in all four settings, with 15, 14, 15, and 15 wins, respectively. Notably, even its lowest average accuracy, 96.36\% in the 2-shot setting, remains higher than the best average accuracy achieved by any baseline at any of the reported support sizes.

The effect of increasing the support size is nevertheless dataset-dependent. Highly separable datasets such as BME, Plane, and GunPointOldVersusYoung already reach 100\% accuracy with OSSE-LSTM using only two examples per class and remain essentially saturated as additional support samples are introduced. In contrast, datasets with greater intra-class variability show greater benefit from additional support. For example, OSSE-LSTM improves on ArrowHead from 81.95\% at 2-shot to 85.45\% at 6-shot, while its accuracy on Wine increases from 83.22\% to 86.31\% over the same range. These improvements are not necessarily monotonic at the dataset level, indicating that additional support samples can introduce intra-class variation as well as useful class information.

At the aggregate level, however, OSSE-LSTM is remarkably insensitive to the support-set size: its average accuracy varies by only 0.36 percentage points across the four settings. This behavior contrasts with several baselines. InceptionTime decreases from 90.26\% at 2-shot to 78.12\% at 8-shot, and MiniRocket decreases from 88.79\% to 78.88\%, whereas TS2Vec improves from 76.72\% to 85.73\% as more support examples become available. LSTM-FCN and TapNet are comparatively stable, but their average accuracies remain below that of OSSE-LSTM in every setting. Overall, the results in Tables I and II indicate that OSSE-LSTM can form reliable class prototypes from very few labeled examples while maintaining robust performance as the support set grows.

\subsection{Ablation Study}

Table~\ref{tab:ablation_study} evaluates the contribution of SE recalibration and the BiLSTM branch on a 14-dataset subset, comparing the base OS-CNN, OS-CNN+SE, OS-CNN+BiLSTM, and the full OSSE-LSTM.

\begin{table*}[htbp]
    \centering
    \caption{Ablation on SE recalibration and BiLSTM temporal modeling, reported on a 14-dataset subset of the UCR benchmarks. Average accuracy and best-count summarize the overall trend; best-count treats ties as wins.}
    \label{tab:ablation_study}
    \resizebox{0.84\textwidth}{!}{
    \begin{tabular}{l cccc cccc}
        \toprule
        \multirow{2}{*}{\textbf{Dataset}} 
        & \multicolumn{4}{c}{\textbf{Ratio = 40\%}} 
        & \multicolumn{4}{c}{\textbf{Ratio = 90\%}} \\
        \cmidrule(lr){2-5} \cmidrule(lr){6-9}
        & \textbf{OS-CNN} & \textbf{+ SE} & \textbf{+ LSTM} & \textbf{+ SE + LSTM} 
        & \textbf{OS-CNN} & \textbf{+ SE} & \textbf{+ LSTM} & \textbf{+ SE + LSTM} \\
        \midrule
        ArrowHead                 & \textbf{71.4} & 67.1 & 64.8 & 65.7 & 76.7 & 77.5 & 79.9 & \textbf{80.9} \\
        Chinatown                 & 96.3 & 93.5 & 96.9 & \textbf{97.1} & 98.0 & 98.4 & 98.3 & \textbf{99.0} \\
        ECG200                    & 81.0 & \textbf{81.1} & 81.0 & 80.4 & 88.1 & \textbf{91.1} & 88.0 & 88.7 \\
        GunPointAgeSpan           & 97.3 & 98.0 & 97.3 & \textbf{98.2} & 99.5 & \textbf{99.7} & 99.6 & \textbf{99.7} \\
        GunPointMaleVersusFemale  & \textbf{100.0} & 99.9 & \textbf{100.0} & \textbf{100.0} & \textbf{100.0} & 99.9 & \textbf{100.0} & \textbf{100.0} \\
        GunPointOldVersusYoung    & \textbf{100.0} & \textbf{100.0} & \textbf{100.0} & \textbf{100.0} & \textbf{100.0} & \textbf{100.0} & \textbf{100.0} & \textbf{100.0} \\
        MoteStrain                & \textbf{87.7} & 86.3 & 87.5 & 85.6 & \textbf{92.3} & 90.6 & 91.3 & 90.7 \\
        SonyAIBORobotSurface1     & 91.2 & 89.8 & \textbf{94.9} & 94.0 & 95.9 & 95.8 & 95.8 & \textbf{97.7} \\
        SonyAIBORobotSurface2     & 93.8 & 92.8 & 93.8 & \textbf{95.5} & 96.1 & 95.2 & 96.0 & \textbf{97.6} \\
        SyntheticControl          & 98.9 & 99.5 & 98.9 & \textbf{99.7} & 99.4 & 99.4 & 99.4 & \textbf{99.9} \\
        ToeSegmentation1          & 91.2 & \textbf{91.6} & 91.3 & 91.5 & 95.4 & 94.4 & 95.2 & \textbf{95.7} \\
        TwoLeadECG                & 97.4 & 96.5 & \textbf{99.6} & 99.3 & 99.7 & 99.7 & 99.7 & \textbf{99.9} \\
        UMD                       & 96.9 & 95.1 & \textbf{98.9} & 97.1 & \textbf{99.3} & \textbf{99.3} & \textbf{99.3} & \textbf{99.3} \\
        Wine                      & 78.7 & \textbf{79.7} & 78.1 & 77.6 & 82.0 & \textbf{83.4} & 82.1 & 82.9 \\
        \midrule
        \textbf{Average Accuracy} & 91.56 & 90.78 & \textbf{91.64} & 91.55 & 94.46 & 94.60 & 94.61 & \textbf{95.14} \\
        \textbf{Best-count}       & 4 & 4 & 5 & \textbf{6} & 4 & 5 & 3 & \textbf{11} \\
        \bottomrule
    \end{tabular}
    }
\end{table*}

We can see that components contribute differently across datasets. SE recalibration helps when only a subset of convolutional channels is informative (\textit{ECG200}, \textit{Wine}) but can suppress useful responses when discriminative local patterns are already clear (\textit{ArrowHead}, \textit{MoteStrain}). The BiLSTM branch is most useful where temporal ordering or phase-sensitive evolution contributes to the class boundary (\textit{TwoLeadECG}, \textit{SonyAIBORobotSurface1}). In the low-data regime, the full OSSE-LSTM is not always superior because the recurrent branch adds parameters and can increase support-set sensitivity. At $\rho=0.9$, however, the full model achieves the best average accuracy and the highest best-count, i.e., 11 counts, supporting the dual-stream design as a robust default while showing that different dataset morphologies may favor different components.

\subsection{Interpretability Analysis}

\textbf{ECG200 dataset:} contains two heartbeat categories: a normal heartbeat and a
myocardial-infarction event. Because both classes represent one cardiac
cycle, they share a common overall structure. Their distinction is therefore
expected to depend on the morphology, duration, and relative magnitude of
the principal cardiac deflections.

\begin{figure*}[!h]
    \centering
    \begin{subfigure}[t]{0.48\linewidth}
        \centering
        \includegraphics[width=\linewidth]
        {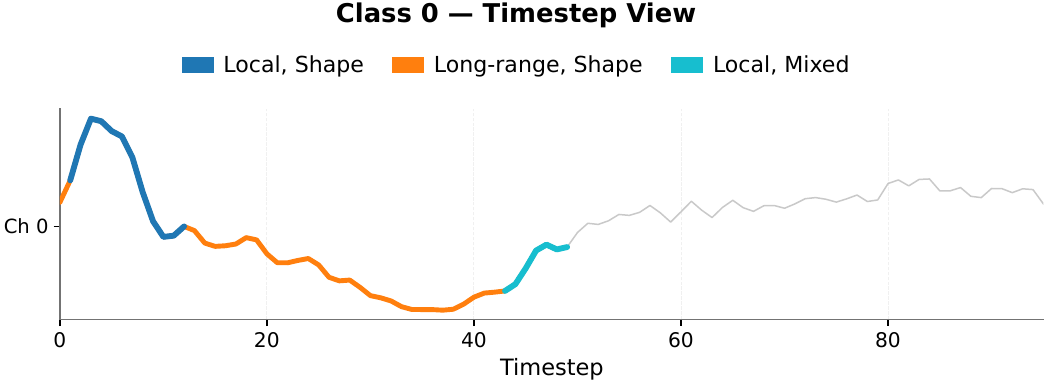}
        \caption{Normal heartbeat}
    \end{subfigure}
    \hfill
    \begin{subfigure}[t]{0.48\linewidth}
        \centering
        \includegraphics[width=\linewidth]
        {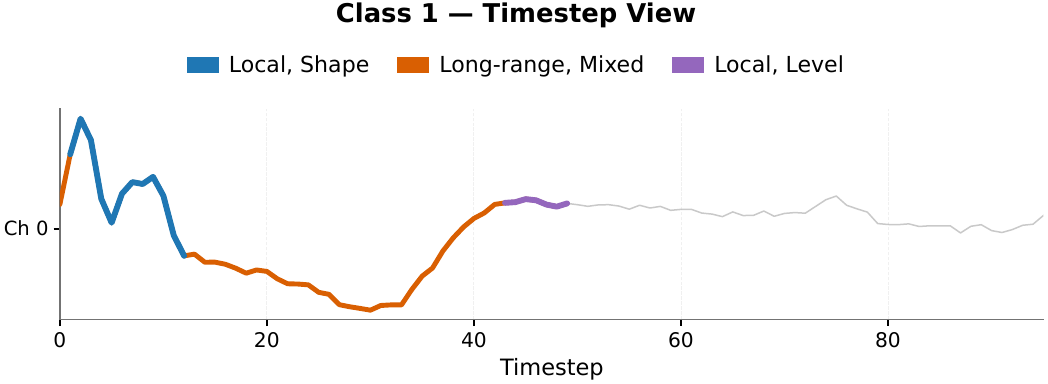}
        \caption{Myocardial-infarction event}
    \end{subfigure}
    \caption{OSSE-LSTM interpretation results on ECG200.}
    \label{fig:appendix_ecg200}
\end{figure*}

\begin{figure*}[!h]
    \centering
    \begin{subfigure}[t]{0.48\linewidth}
        \centering
        \includegraphics[width=\linewidth]
        {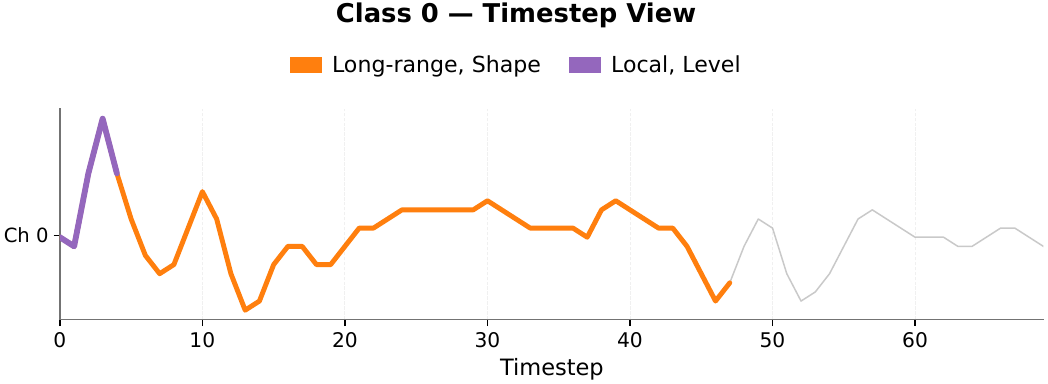}
        \caption{Cement}
    \end{subfigure}
    \hfill
    \begin{subfigure}[t]{0.48\linewidth}
        \centering
        \includegraphics[width=\linewidth]
        {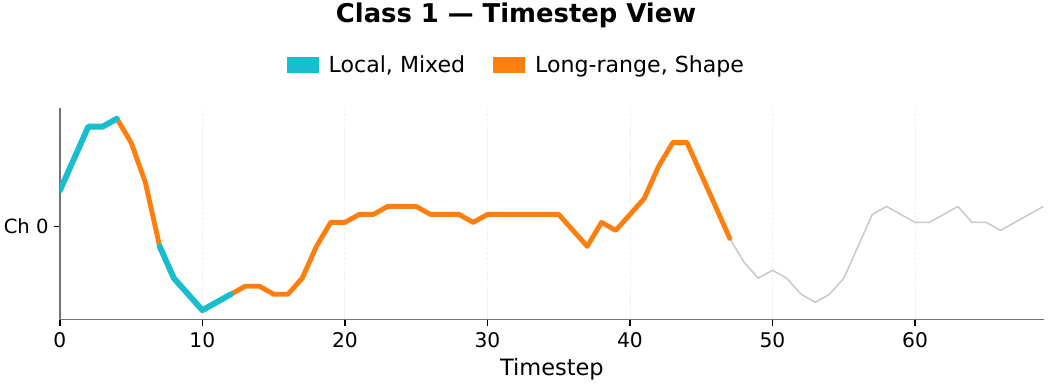}
        \caption{Carpet}
    \end{subfigure}
    \caption{OSSE-LSTM interpretation results on SonyAIBORobotSurface1.}
    \label{fig:appendix_sony}
\end{figure*}

As shown in Fig. \ref{fig:appendix_ecg200}, for the normal heartbeat, OSSE-LSTM emphasizes the initial sharp peak, the
following negative deflection, and the gradual recovery toward the baseline.
The initial peak is represented by a local shape-based region, whereas the
subsequent descent and recovery are captured by a longer shape-based
interval. A shorter mixed region marks the final transition toward the
stable tail.

For the myocardial-infarction event, OSSE-LSTM again focuses on the early
deflections and the main recovery phase, but the selected boundaries and
evidence types differ. The initial oscillatory morphology is represented by
a local shapelet, while the descent into the trough and the steep return are
captured by a long-range mixed region. A short level-sensitive interval near
the end of the recovery indicates that the absolute signal level also
contributes to the distinction.

These explanations are physiologically plausible because the two categories
share the same basic heartbeat structure. OSSE-LSTM does not assign importance
to unrelated positions. Instead, it compares corresponding cardiac phases
and identifies class-dependent changes in peak morphology, trough depth, and
recovery behavior.

\textbf{SonyAIBORobotSurface1 dataset:} contains the \(x\)-axis accelerometer measurements of a Sony AIBO robot walking on cement and carpet. The surface affects the magnitude of each impact, the oscillation following foot contact, and the rate at which the mechanical response is damped.

As shown in Fig. \ref{fig:appendix_sony}, for cement, OSSE-LSTM first identifies a short level-sensitive interval around
the initial high-amplitude impulse. A long-range shapelet then covers the
successive peaks and troughs produced after the initial contact. This
suggests that the decision uses both the strength of the first impact and
the extended vibration response of the robot.

For carpet, a local mixed region captures the initial rise and the rapid
descent into a deep trough. A subsequent long-range shapelet follows the
recovery and later oscillations. Compared with cement, the selected evidence
places greater emphasis on the transition from impact to the first trough
and on the following damped response.

These explanations are consistent with the mechanical difference between a
hard and a soft surface. Cement produces a stronger and less damped response,
whereas carpet modifies both impact magnitude and the evolution of the
following vibration. OSSE-LSTM therefore selects impact and damping structures
rather than arbitrary high-amplitude positions.

\subsection{Attribution-Guided Test-Time Prototype Refinement}

Finally, we evaluate whether C-IG attribution maps can guide test-time prototype refinement. We compare two strategies: random-noise TTA, in which support samples are perturbed without attribution guidance, and C-IG-guided TTA, in which perturbations are applied mainly to low-attribution regions while high-attribution regions are preserved. For each dataset, both variants use the same trained checkpoint but are evaluated in separate runs. Because episodic sampling causes slight variation in the standard accuracy, each TTA result is compared against the standard baseline of the same run.

\begin{table}[htbp]
    \centering
    \caption{Random-noise TTA vs.\ C-IG-guided TTA. Boost is computed relative to the standard accuracy of the same run.}
    \label{tab:cig_tta_results}
    \resizebox{\linewidth}{!}{
    \begin{tabular}{lcccccc}
        \toprule
        \multirow{2}{*}{\textbf{Dataset}} 
        & \multicolumn{3}{c}{\textbf{Random Noise TTA}} 
        & \multicolumn{3}{c}{\textbf{C-IG TTA}} \\
        \cmidrule(lr){2-4} \cmidrule(lr){5-7}
        & \textbf{Std.} & \textbf{TTA} & \textbf{Boost}
        & \textbf{Std.} & \textbf{TTA} & \textbf{Boost} \\
        \midrule
        ArrowHead 
        & $78.89\pm1.35$ & $73.56\pm1.40$ & $-5.33$
        & $79.86\pm1.30$ & $80.58\pm1.26$ & $+0.72$ \\
        ECG200
        & $91.00\pm1.20$ & $90.58\pm1.28$ & $-0.42$
        & $88.67\pm1.39$ & $88.92\pm1.38$ & $+0.25$ \\
        ItalyPowerDemand
        & $94.83\pm0.91$ & $94.79\pm0.91$ & $-0.04$
        & $94.92\pm1.01$ & $95.04\pm0.98$ & $+0.12$ \\
        Wine
        & $83.92\pm1.49$ & $60.29\pm1.46$ & $-23.63$
        & $78.12\pm1.64$ & $78.50\pm1.60$ & $+0.38$ \\
        \bottomrule
    \end{tabular}
    }
\end{table}

Random-noise TTA degrades accuracy on all four datasets, severely so on \textit{Wine}, confirming that unguided perturbation is risky in few-shot TSC: it may corrupt the localized evidence needed to form reliable prototypes. C-IG-guided TTA avoids this degradation and provides small but consistent gains. The largest gain appears on \textit{ArrowHead}, whose class differences are tied to local shape morphology and whose standard baseline has not saturated. Gains on \textit{ECG200} and \textit{ItalyPowerDemand} are smaller because their standard prototypes are already relatively stable. On \textit{Wine}, attribution-guided masking prevents the catastrophic degradation observed under random noise.

These results indicate that C-IG maps are more than visual artifacts: preserving high-attribution regions while perturbing low-attribution regions can maintain or improve prototype quality. The effect is dataset-dependent, however, and C-IG-guided TTA should be regarded as an auxiliary refinement strategy rather than a universal improvement. %Future work could adapt the perturbation scale to prototype confidence, signal smoothness, and attribution concentration.

\section{Conclusion}
\label{sec:conclusion}

We introduced \textbf{Dual-Stream OSSE-LSTM}, an episodic metric-learning
framework for label-efficient time series classification, where a fixed set of
classes must be recognized from only a few labeled examples each. It integrates
Omni-Scale convolution for multi-scale local features with BiLSTM modeling for
bidirectional temporal context, applying stream-wise normalization before
prototype-based classification.
On 19 univariate UCR datasets against six baselines, OSSE-LSTM attains the highest
average accuracy and win count at every support size from $K=2$ to $K=8$, staying
within a $0.36$-point band ($96.36$--$96.72\%$). Its weakest configuration
exceeding the best result any baseline reaches at any $K$ ($93.99\%$), while
convolutional baselines lose up to $12$ points as $K$ grows. %Two labeled examples per class are therefore sufficient, and further annotation yields almost no additional accuracy.

We also proposed \textbf{Counterfactual Integrated Gradients (C-IG)}, which
interprets prototype-based decisions by attributing the target-opponent prototype
margin and, applied to test-time refinement, gives small but consistent gains
where unguided noise augmentation degrades accuracy.

\bibliographystyle{IEEEtran}
\bibliography{references}

% that's all folks
\end{document}